%% file: paper.tex
\documentclass[10pt,twocolumn,letterpaper,final]{article}

\usepackage{cvpr}
\usepackage{times}
\usepackage{epsfig}
\usepackage{graphicx}
\usepackage{amsmath}
\usepackage{amssymb}
\usepackage{subfig}
\usepackage{enumitem}
\usepackage{ctable}
\usepackage{booktabs}
\usepackage{arydshln}

\newcommand{\norm}[1]{\left\lVert#1\right\rVert}

\newcommand{\out}[1]{}
\newcommand{\mypar}[1]{\noindent\textbf{#1}}

\usepackage[pagebackref=true,breaklinks=true,letterpaper=true,colorlinks,bookmarks=false]{hyperref}

\cvprfinalcopy 

\def\cvprPaperID{3533} 

\ifcvprfinal\fi

\begin{document}

\title{Rules of the Road:\\Predicting Driving Behavior with a Convolutional Model of Semantic Interactions}

\author{Joey Hong\thanks{Work done while at Zoox.  The authors would also like to thank and acknowledge Kai Wang (kai@zoox.com) for his work on this project. Kai has been instrumental in coordinating the final version of the paper and preparing the dataset for release.}\\
Caltech\\
{\tt\small jhhong@caltech.edu}
\and
Benjamin Sapp\footnotemark[1]\\
{\tt\small benjamin.sapp@gmail.com}
\and
James Philbin\\
Zoox\\
{\tt\small james@zoox.com}
}

\maketitle

\input{abstract}
\input{intro}

\input{relwork}

\input{method}

\input{experiments}

{\small
\bibliographystyle{ieee}
\bibliography{bib}
}

\end{document}

%% file: abstract.tex
\begin{abstract}
We focus on the problem of predicting future states of entities in complex, real-world driving scenarios.  Previous research has used low-level signals to predict short time horizons, and has not addressed how to leverage key assets relied upon heavily by industry self-driving systems: (1) large 3D perception efforts which provide highly accurate 3D states of agents with rich attributes, and (2) detailed and accurate semantic maps of the environment (lanes, traffic lights, crosswalks, etc). We present a unified representation which encodes such high-level semantic information in a spatial grid, allowing the use of deep convolutional models to fuse complex scene context. This enables learning entity-entity and entity-environment interactions with simple, feed-forward computations in each timestep within an overall temporal model of an agent's behavior.  We propose different ways of modelling the future as a {\em distribution} over future states using standard supervised learning. We introduce a novel dataset providing industry-grade rich perception and semantic inputs, and empirically show we can effectively learn fundamentals of driving behavior.
\end{abstract}

%% file: intro.tex
\section{Introduction}

\begin{figure}[t]
\begin{center}
\includegraphics[width=0.99\linewidth]{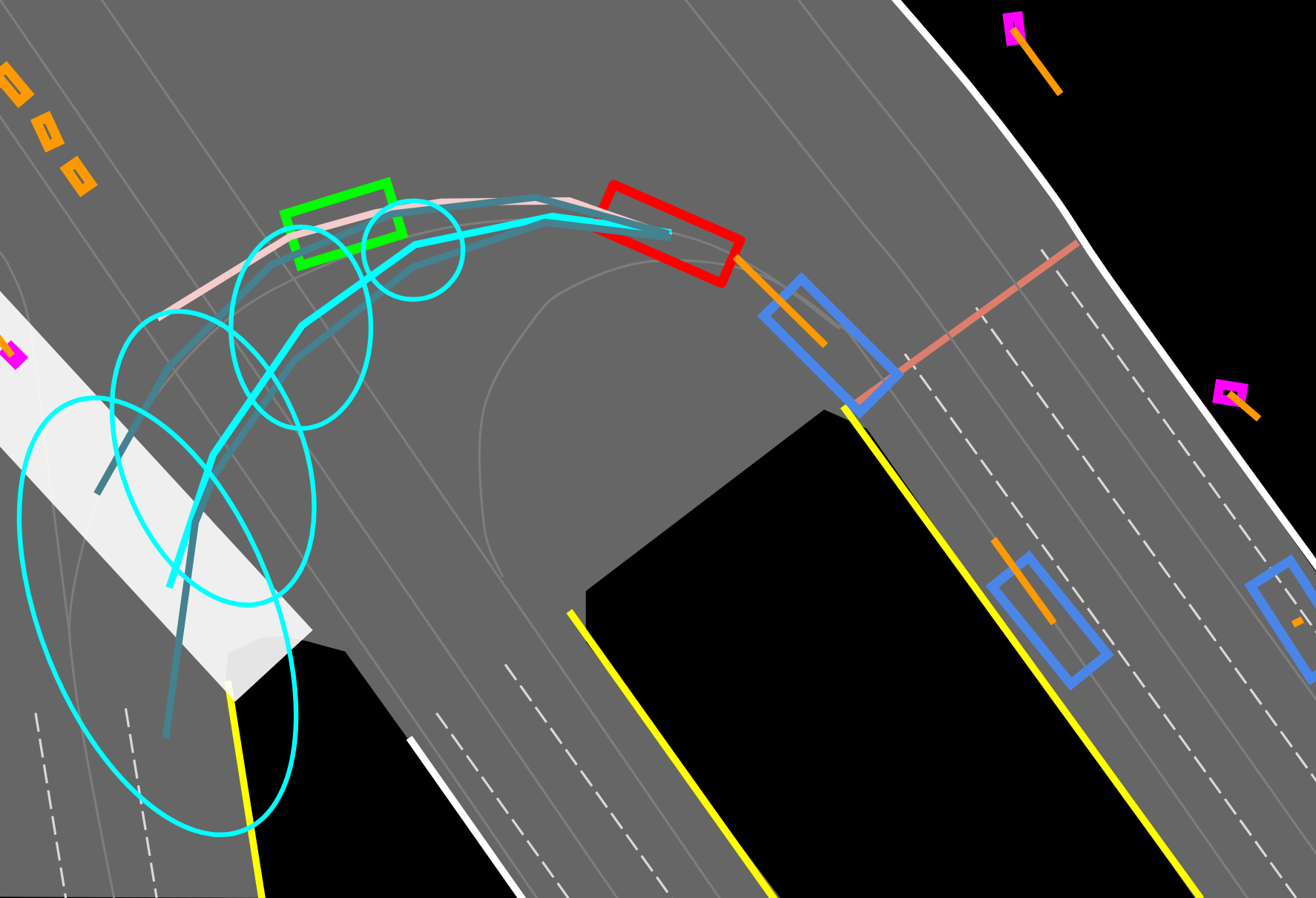}

\end{center}
\captionsetup{singlelinecheck=off}
\caption{Entity future state prediction task on a top-down scene:
A target entity of interest is shown in red, with a real future trajectory shown in pink.  The most likely predicted trajectory is shown in cyan, with alternate trajectories shown in green.  Uncertainty ellipses showing 1 standard deviation of uncertainty are displayed for the most likely trajectory only.  Other entities are rendered in magenta (pedestrians), blue (vehicles) and orange (bicycles).  The ego vehicle which captured the scene is shown in green. Velocities are shown as orange lines scaled proportional to $1 m/s$. Examples of underlying semantic map information shown are lane lines, crosswalks and stop lines.
}
\label{fig:snapshot}
\end{figure}

A crucial component of real-world robotic systems is predicting future states of other actors in the environment.  In the general setting, other actors' intents are unobserved, thus the challenge is to produce a likely distribution over possible futures, given current and past observations. A motivating application for this work is a self-driving robot operating in an unconstrained urban environment; one of the most impactful yet challenging real-world robotics applications today.  It requires a deep understanding of the semantics of the static and dynamic environment, including understanding traffic laws, unspecified driving conventions, and interactions between human and robot actors.

While there is a large amount of research dedicated to real-world perception in 
this domain~\cite{Kitti,Faf,mousavian20173d,BaiduBEV,VoxelNet,PointNet,CarFusion,Bullinger_2018_ECCV}, there is a surprising lack of work on entity state prediction in the same domain (see Section~\ref{sec:relwork}), which we attribute to two main causes:

One, most previous research~\cite{Faf,Desire} takes as input only raw sensor information (a combination of camera, lidar or radar).  Thus the research effort by necessity requires a heavy emphasis on extracting high-level representations of entities.  In contrast, in real-world systems in industry, low-level perception systems provide entity states and attributes from sensor data via detection and tracking in 2d and 3d. These systems are have matured over the past few years and have high fidelity output in the common case.

Two, publicly available datasets for learning and evaluating state prediction 
models are inadequately small and/or unrealistic.  A good prediction dataset 
should include a diverse set of real-world locations and a large number of 
unique agent 3d tracks over meaningful time intervals.  These criteria are 
necessary to develop models which can generalize to new scenarios and leverage 
past behavior to make meaningful future predictions out to 5 seconds or more. A final omission in previous prediction research is a key asset relied upon heavily by industry self-driving robots: semantic maps of the driving environment, as shown in Fig.~\ref{fig:snapshot}.

In this paper, we introduce a vehicle prediction dataset which is significantly 
richer and larger than existing datasets---9,659 unique vehicles in 83,880 prediction scenarios (173 hours), in 88 physically-distinct locations---and includes semantic map information (see 
Figure~\ref{fig:snapshot}). 

We propose a model which encodes a history of world state (both static and dynamic) and semantic map information in a unified, top-down spatial grid.  This allows us to use a deep convolutional architecture to model entity dynamics, entity interactions, and scene context jointly.

An additional important contribution of this work is directly predicting {\em distributions} of future states, rather than a single point estimate at each future timestep.  Representing multimodal uncertainty is crucial in real-world planning for driving, which must consider vehicles taking different possible trajectories, or assessing the expected risk of collision within some spatial extent.

We explore a variety of parametric and non-parametric output distribution representations, and show strong performance on predicting vehicle behavior up to 5 seconds in the future. We demonstrate that our model leverages road information and other agents' state to improve predicted behavior performance.

%% file: relwork.tex
\section{Related Work}
\label{sec:relwork}

This paper focuses on predicting future distributions of entity state.  This requires implicitly or explicitly modeling entity intent, dynamics, and interactions, as well as incorporating semantic environmental context.

\mypar{Activity/motion forecasting}
Early work by Kitani \etal \cite{Kitani} formulated this setting as a Partially-Observable Markov Decision Process, and cast the solution as recovering a policy via Inverse Optimal Control (IOC).  More recently, Rhinehart \etal~\cite{R2p2} also learn one-step control policy distributions within a reinforcement learning framework, specifically trained (via symmetric KL-divergence loss) to generate a set of trajectories with a balance between the notions of diversity and precision. DESIRE~\cite{Desire} also generates trajectories out to $4s$ by iteratively rolling out a one-step policy, via sampling in a Conditional Variational Auto Encoder / Decoder module trained end-to-end in conjunction with a trajectory-ranking module. Other work also concentrates on multimodal distribution modeling~\cite{Multi-human, Trajectory-gmm}, but do so outside the self-driving domain. 

Fast and Furious~\cite{Faf} and follow-on work IntentNet~\cite{IntentNet} take a standard supervised regression approach; so-called {\em behavior cloning} in the RL literature. These works emphasize joint detection (from lidar), tracking and motion forecasting in one model.  The recently published IntentNet uses a semantic road map information as an input. It proposes single trajectories per entity out to $3s$.  

Relatedly, ChaufferNet~\cite{Chauffernet} fuses high-level agent information (as we do) with a road map. This work focuses on the robot planning setting where the intent is known and fed as input.  The model employs behavior cloning to regress the best motion plan, and employs a number of domain-specific losses to encourage road rule-following and collision avoidance. Other notable behavior-cloning approaches in the autonomous vehicle industry are~\cite{ALVINN, nvidia-sdc, AutoX}.

Another line of work attempts forecasting from ego-centric viewpoints (moving camera frame), either of the ego-entity~\cite{ego-pred2} or of other entities~\cite{ego-pred}.  This introduces the additional challenge of needing to recover the ego-position and/or velocity, a problem these works address.
There is a wealth of other forecasting work which we don't review here, limiting this discussion to the state prediction in multi-agent environments, although there are connections to unsupervised video prediction (\eg~\cite{VideoPred}) and activity prediction (\eg~\cite{ActivityPred}).

\mypar{Modeling entity interactions} Much of the above research models interactions implicitly by encoding them as surrounding dynamic context. Other works explicitly model interactions: SocialLSTM~\cite{Sociallstm} pools hidden temporal state between entity models,~\cite{bagautdinov2017social} is one example modeling explicit graph structure over entities to infer semantic actions, and~\cite{Kooij2014ContextBasedPP} is an earlier work posing the structure as a Bayesian network.

\mypar{Vehicle Forecasting Datasets}
We evaluate on vehicles in this work, and propose a new dataset of roughly 80K examples / 170 hours of data.  Related datasets are Kitti~\cite{Kitti}, primarily a detection and tracking benchmark, which has about 50 examples / 10 minutes of data; IntentNet's dataset~\cite{IntentNet} (unpublished) is roughly 5000 examples  / 35 hours, and CaliForecasting~\cite{R2p2} (as yet unreleased) is 10K examples / 1.5 hours.

\mypar{Contrast with our method} 
Our method is, to our knowledge, the only end-to-end method that both (1) encodes semantic scene context and entity interactions from a mature perception stack, as well as (2) predicts multimodal future state distributions. 

Most similar to our work, the recent IntentNet~\cite{IntentNet} and ChaufferNet~\cite{Chauffernet} papers propose encoding static and dynamic scene context in a top-down rasterized grid. A main difference in our work is that we explicitly model multimodal distributions for other entities, rather than regress single trajectories. This is an important and challenging task. 

DESIRE~\cite{Desire} and R2P2~\cite{R2p2} address multimodality, but both do so via 1-step stochastic policies, in contrast to ours which directly predicts a time sequence of multimodal distributions. Such policy-based methods require both future roll-out and sampling to obtain a set of possible trajectories, which has computational trade-offs to our one-shot feed-forward approach.  Also, knowing how many samples are required to be confident in your empirical distribution is a hard problem, and depends on the scenario.

%% file: method.tex
\begin{figure*}[t]
\begin{center}
\includegraphics[width=0.99\linewidth]{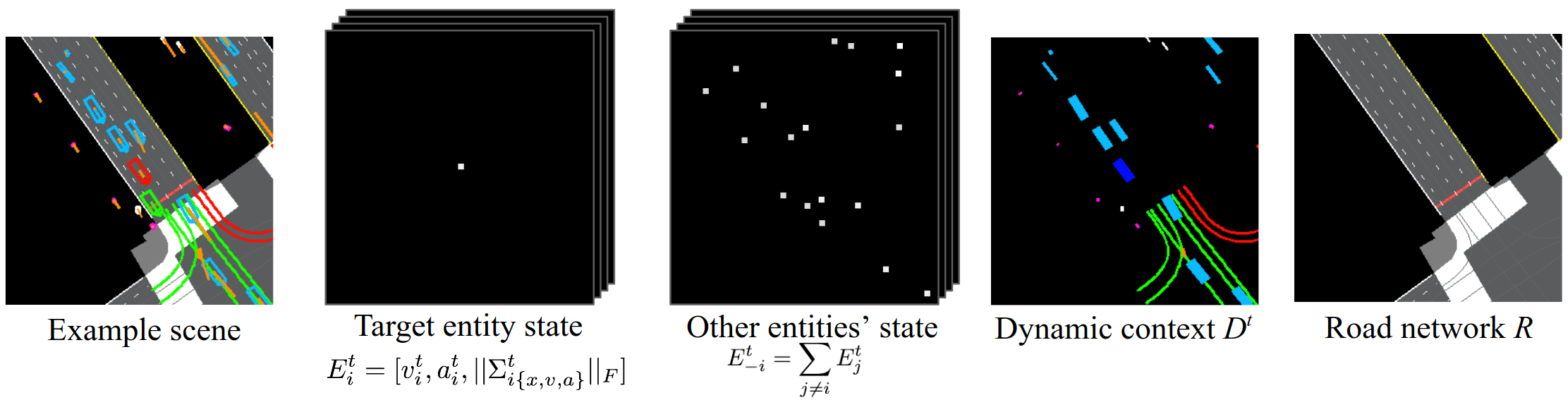}
\end{center}
\vspace*{-0.5cm}
\caption{Entity and world context representation.  For an example scene (visualized left-most), the world is represented with the tensors shown, as described in the text.}
\vspace*{-0.2cm}
\label{fig:feature_extract}%
\end{figure*}

\section{Method}

Our method consists of (1) a novel input representation of an entity and it's 
surrounding world context, (2) a neural network mapping such past and present 
world representation to future behaviors, and (3) one of several possible 
output representations of future behavior amenable to integration into a robot 
planning system.  Note our model is an {\em entity-centric} model, meaning it 
models a single ``target entity'', but takes all other world context, 
including other entities,  into account\footnote{At run time, straightforward 
extensions can enable  inference for all entities in the scene efficiently, 
with most of the  computation re-used for each agent (since each differs only 
in a translation  of the input). }.

\subsection{Input representation: modeling the world as a convolutional grid}

A complete model of trajectory prediction requires not only past history of 
the target entity, but also dynamics of other entities and semantic scene 
context.

\mypar{Road network representation}
\label{sec:roadgraph}
We have access to road network data which includes lane and junction extent 
and connectivity, as well as other relevant features necessary for driving: 
crosswalks, traffic light-lane permissibility, and stop and yield lines\footnote{This type of content is freely available as open source data, \eg at \url{www.openstreetmap.org}, but we use a proprietary source in this work.}.  We map this information to geometric primitives, and render it in a top-down grid representation as an RGB image with unique colors corresponding to each element type.  

This top-down grid establishes the common coordinate 
space with which we register all additional features.  Note that through 
rendering, we lose the true graph structure of the road network, leaving it as 
a modeling challenge to learn valid road rules like legal traffic direction, and valid paths through a  junction.
We denote the rendered tensor of static road information $R$ of size $W \times H \times 3$.  Traffic light information is added to a tensor of perception information per timestep described below.

\mypar{Entity representation}
We assume access to a black-box perception module which maps from low-level sensor information to 3D tracked entities\footnote{This is a reasonable assumption for large companies where such modules are relatively mature.}.
For each timestep $t$, we have measured quantities for each tracked entity $i$
including 2D position $x^t_i$, velocity $v_i^t$, and acceleration $a_i^t$.  Our 
perception module also gives us state estimation uncertainty in the form 
of covariance matrices, and we include this information in our representation 
via covariance norms $||\Sigma^t_{i\{x,v,a\}}||_F$.  All feature dimensions are scaled by an estimate of their $99^{th}$ percentile magnitude to have comparable dynamic ranges near $[-1,1]$.

We form a tensor for the target entity $i$ at any timestep $t$ denoted $E_i^t
$, that has a channel for each state dimension above, encoding the scalar at the  center of the entity position, which is in spatial correspondence with road graph tensor $R$. To model entity 
interactions, we aggregate all other entities in a tensor encoded in a the 
same way: $E_{-i}^t = \sum_{j \neq i} E^t_j$.  These tensors are $W \times H \times 7$.

\mypar{Additional dynamic context}
We encode additional scene context as an RGB image $D^t$ of size $W \
\times H \times 3$.  It contains oriented bounding boxes for all entities in 
the scene colored by class type (one of cyclist, vehicle, pedestrian), to 
encode extents and orientations of objects. It also contains a rendering of 
traffic light permissibility in junctions: we render permitted (green light),  
yield (unprotected), or prohibited (red light ) by masking road connections 
in  a junction that exhibit each permissibility.

\mypar{Temporal modeling}
All inputs at timestep $t$ and target entity $i$ are concatenated (in the third channel dimension) into a tensor:

$$C^t_i  = \Big[E_i^t, E_{-i}^t, D^t, R\Big]$$ which is $W \times H \times 20$. 
See Fig.~\ref{fig:feature_extract} for an illustration. We concatenate all $C^t_i$ over past history along a temporal dimension. We fix the coordinate system for a static $R$ for all timestamps by centering the reference frame at the position of the target entity at the time of prediction.

This top-down representation is simple to augment with additional entity features in the future. For example: vehicle brake lights and turn signals, person pose and gestures, and even audio cues could all be integrated as additional state channel dimensions.

\begin{figure*}%
\begin{center}
\subfloat[One-shot]{{\includegraphics[width=0.48\linewidth,height=4cm]{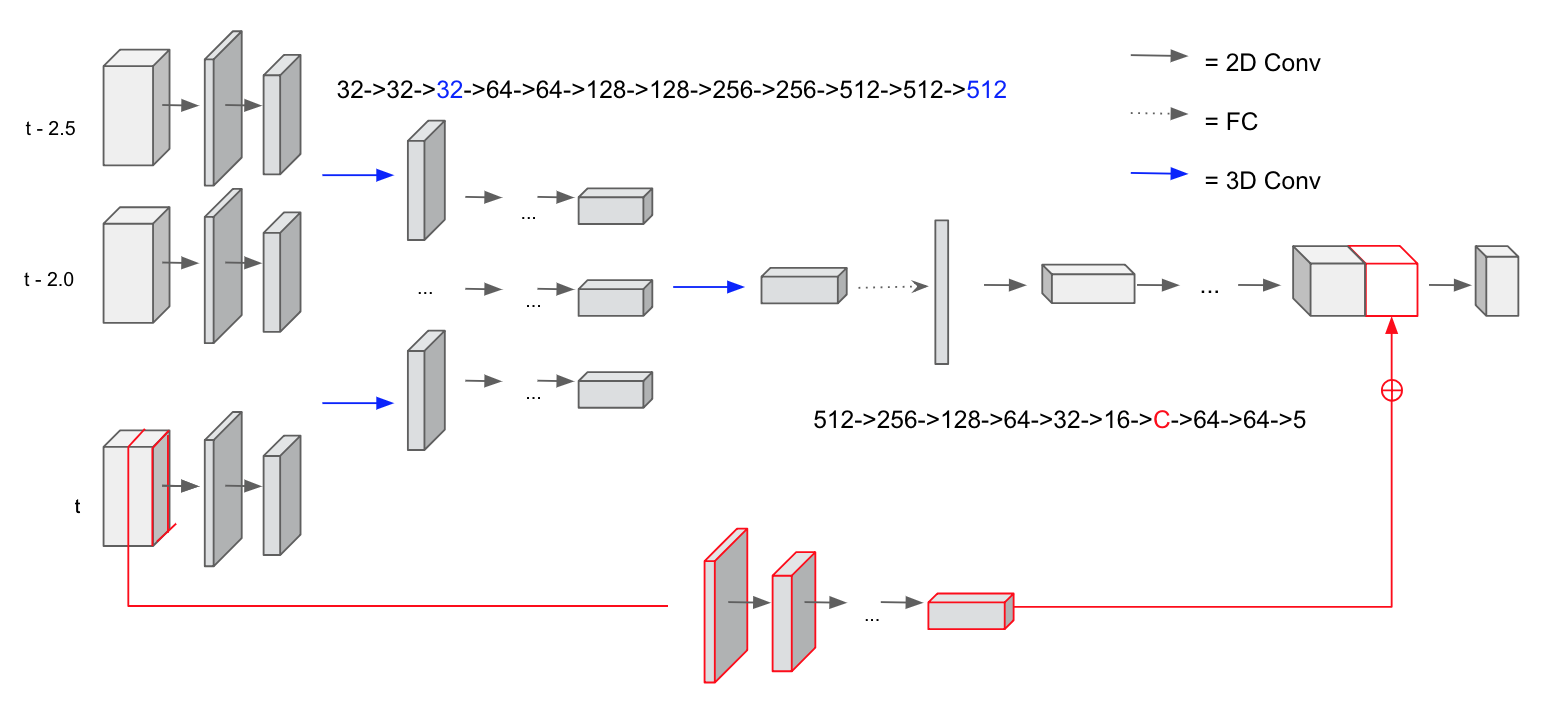} }}%
\quad
\subfloat[with RNN decoder]{{\includegraphics[width=0.48\linewidth,height=4cm]{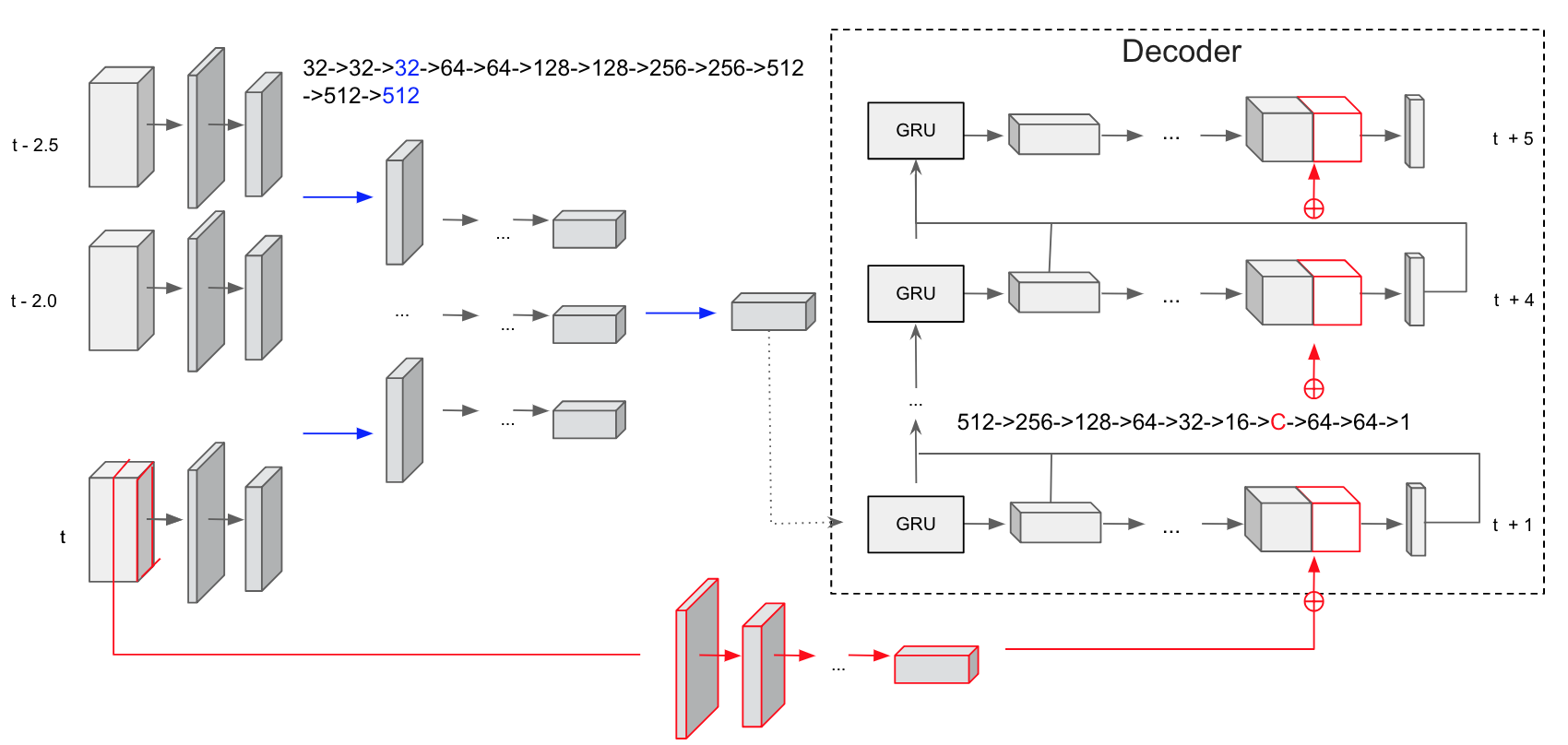} }}%
\caption{Two different network architectures for occupancy grid maps (predicting Gaussian trajectories instead is done by simply replacing the convolutional-transpose network with a fully-connected layer).}%
\label{fig:architectures}%
\end{center}
\end{figure*}

\subsection{Output representation: modeling uncertainty and multiple modes}

We seek to model an entity's future states.  We believe a good output representation must have the following characteristics, which differs from some previous work~\cite{IntentNet, R2p2}. It should be:
(1) {\em A probability distribution} over the 
entity state space at each timestep. The future is inherently uncertain, and a 
single most-likely point estimate isn't sufficient for a safety-critical system.
(2) {\em Multimodal}, as it is  important to cover a diversity of possible 
implicit actions an entity might take (\eg, which way through a junction). 
(3) {\em One-shot}: For efficiency reasons, it is desirable to predict full 
trajectories (more specifically: time sequences of state distributions) without iteratively applying a recurrence step.

We explore these desiderata by proposing a variety of output distribution 
representations.  The problem can be naturally formulated as a sequence-to-sequence generation problem. For a time instant $t$, we observe $X = \{C_{t-\ell + 1}, C_{t-\ell+2}, \ldots , C_{t} \}$ of past scenes, and predict future xy-displacements from time $t$, denoted $Y = 
\{ (x_{t+1}, y_{t + 1}), (x_{t + 2}, y_{t + 2}), \ldots ,(x_{t + m}, y_{t+m}) 
\}$, for a target entity\footnote{We ignore modeling future entity orientation, but we believe it is straightforward to extend the ideas here to 3-dof or 6-dof 
state estimation in future work.}, where $\ell$ and $m$ are the time range 
limits for past observations and future time horizon we consider.  We discuss a variety of approaches to model $P(Y|X)$ next.

\subsection{Parametric regression output with uncertainty}  \out{
Since the actual trajectories of an entity follow a posterior distribution, 
instead of predicting raw displacements we also let the model predict 
uncertainty. By doing this, the model can attenuate the effect of outlier 
trajectories to have a smaller effect on the loss ~\cite{Uncertainty}. 
} 
In this representation, we predict sufficient statistics for a bivariate 
Gaussian for each future position $(x_t, y_t)$: the mean $\mu_t = (\mu_{x, t}, \mu_{y, t
})$, standard deviation $\sigma_t = (\sigma_{x, t}, \sigma_{y, t})$, and 
spatial correlation scalar $\rho_t$.  This has benefits of being a richer and more useful 
output than vanilla regression.  In addition, during learning, the model can 
attenuate the effect of outlier trajectories in the data to have a smaller 
effect on the loss, an observation made in other problem domains~\cite{Uncertainty}.

We train the model to predict log standard deviation $s_t = \log \sigma_t$, as 
it has a larger numerically-stable range for optimization. We learn parameters via 
via maximum likelihood estimation
\begin{equation}
\log P(Y \mid X) = \sum_{t' = t + 1}^{t+m} \log p(x_{t'}, y_{t'} \mid \mu_{t'}, s_{t'}, \rho_{t'}), 
\end{equation}
where $p$ is the density function of a bivariate Gaussian parameterized by $\mathcal{N}(\mu_{t}, \sigma_t, \rho_t)$. 

\mypar{Multi-modal regression with uncertainty}
We can extend this to predict up to $k$ different Gaussian trajectories, indexed by 
$(\mu^i, \hat{s}^i, \rho^i)$, and a weight $w^i$ associated with the 
probability of a trajectory,
$$ P(Y \mid X) = \sum_{i=1}^{k} w^i P(Y \mid \mu^i, \hat{s}^i, \rho^i). $$
However, we run into two major problems with this naive method: (1) exchangeability, and (2) collapse of modes. 

For (1), we see that the output is invariant to permutations, namely $\{(\mu^i, s^i, \rho^i)\} = \{(\mu^{\pi(i)}, s^{\pi(i)}, \rho^{\pi(i)})\}$ for some permutation $\pi$ of $\{1, 2, \dots, k\}$. To illustrate (2), we consider a mixture of two distributions with equal covariance, or $Y \sim \frac{1}{2} \mathcal{N}(y^1, \sigma^2) + \frac{1}{2} \mathcal{N}(y^2, \sigma^2)$. It can be shown that if $y^i$ are not far enough, specifically $\mid y^1 - y^2 \mid \leq 2\sigma$ ~\cite{Bimodal}, then the mixture has a single mode at $\frac{1}{2}(y^1 + y^2)$, and thus can be approximated by a single Gaussian parameterized by $\mathcal{N}(\frac{1}{2}(y^1 + y^2), \sigma^2 + \frac{1}{2}(y^1 - y^2)^2)$. This allows the learning to center the $k$ Gaussians on a single state with large variance (``mode collapse''), with nothing to encourage the modes to be well-separated.

Inspired by Latent Dirichlet Allocation (LDA)~\cite{LDA}, we introduce a 
latent variable $z$ such that $(\mu^i, s^i, \rho^i)$ are identically and 
independently distributed conditioned on $z$,
\begin{equation}
P(Y \mid X) = \sum_{i=1}^K P(z_i \mid X) P(Y \mid \mu^i, s^i, \rho^i, z_i),
\end{equation}
where $(\mu^i, s^i, \rho^i)$ is some fixed function of both the input $X$ and 
latent variable $z$, resolving the mode exchangeability problem. To address mode collapse, we choose $z$ to 
be $k$-dimensional with small $k$, and the model is encouraged through 
the learning procedure to output a diverse set of modes. 

To train such a model, we use a conditional variational autoencoder (CVAE)  approach, and model $P(z|X)$ as a categorical distribution, using a Gumbell-Softmax distribution with the reparameterization trick~\cite{GumbelSoftmax} to sample and backpropagate gradients through $z$. In our experiments, we use $P(z|X)$ as our discrete, $k$-Gaussians mixture distribution over state at future timesteps, and refer to this method in our experiments (loosely) as ``GMM-CVAE''.

\out{
, but give a categorical distribution to 
latent variable $z$, instead of the continuous Gaussian $\mathcal{N}(0, I)$ 
typically used on similar tasks. Like with CVAE, we use multiple networks: a 
recognition network $q_{\phi}(z_i \mid X, Y)$, a prior network $p_{\upsilon}(
z_i \mid X)$, and a generative network $p_{\theta}(Y \mid X, z_i)$. 

In train time, we approximate $P(z_i \mid X)$ with a recognition network $q_{\phi}(z_i \mid X, Y)$. The recognition network is given the entire context, including $Y$, and is trained to maximize the log-likelihood of reconstructing $Y$. We do this by sampling $z_i \sim q_{\phi}(z_i \mid X, Y)$, and feeding $z_i$ into the generative network $p_{\theta}(Y \mid X, z_i)$, and backpropogating through it. We use the Gumbell-Softmax reparameterization trick ~\cite{GumbelSoftmax} to sample categorical variable $z$ from some class probability output to accomplish this.

In test time, we use prior network $p_{\upsilon}(z_i \mid X)$ to sample $z_i$ instead, since we don't have access to $Y$. We train the prior network by minimizing the KL-divergence between its class probability outputs, and that of $q_{\phi}(z_i \mid X, Y)$. In summary, we train on the new objective,
\begin{equation}
\begin{split}
\mathcal{L}(\theta, \phi, \upsilon) \,= &-\mathrm{KL}(q_{\phi}(z_i \mid X, Y) \, \| \, p_{\upsilon}(z_i \mid X)) \, + \\
&\qquad \log p_{\theta}(Y \mid X, z_i),
\end{split}
\end{equation}
The weights of predicted trajectories are simply the output of prior network $p_{\upsilon}(z_i \mid X)$.

// out: 
} 

\subsection{Non-parametric output distributions}
\label{sec:gm}
As an alternative to parametric forms, we consider occupancy grid maps~\cite{Gridmap} as an output representation, where there is an output grid for each future modeled timestep, and each grid location holds the probability of the corresponding output state. 

This representation trades off the compactness of the parametric distributions discussed above with arbitrary expressiveness: non-parametric distributions can capture non-elliptical uncertainty and a variable number of modes at each timestep.  Furthermore, they offer alternative ways of integrating into robot 
planning systems (simple to combine with other non-parametric distributions; 
fast approximate integration for, \eg, collision risk calculation).  

We discretize any future state to 2D grid coordinates $(i_t, j_t)$. We train a model to maximize log-likelihood on the predicted grid maps $g_t[i,j] \equiv P(Y=(i_t, j_t)|X)$, which are discrete distributions over the discretized state space at each timestep.  Thus the training loss is a sum of cross-entropy losses for $t' \in [t+1, t+m]$.

\mypar{Diverse Trajectory Sampling:} While occupancy maps have intrinsic representational benefits discussed above, it is still useful in many planning applications to extract a discrete set of trajectories.  We discuss an optimization framework to obtain a variable number of trajectories derived from a future occupancy map, with the ability to impose hard and soft constraints on geometric plausibility and diversity / coverage of the trajectory set. 

Let $\xi_t = (i_t, j_t)$ be a sampled discretized state at time $t$, and $\xi = \{\xi_{t+1}, \ldots, \xi_{t+m} \}$ be a sampled trajectory. We define a pairwise-structured score for a sampled trajectory as:
\begin{equation}
s(\xi) = \sum_{t' = t+1}^{t+m} \log P(Y=\xi_{t'} | X) - \lambda \cdot \phi(\xi_{t'}, \xi_{t'-1}).
\end{equation}
where $\phi(\cdot, \cdot)$ can be an arbitrary score function of the compatibility of consecutive states.  We designed a hard-soft constraint cost
$$\phi(\xi_{t}, \xi_{t-1}) = \begin{cases} \norm{ \xi_{t} - v(\xi_{t - 1}) }_2 ^ 2 &\text{if } \norm{\cdot}_{\infty} \leq 5 \\
\infty &\text{otherwise,} \end{cases} $$
where $\lambda = 0.1$ was set by hand in our experiments and ${v}(\xi_{t})$ is the state transitioned to under a constant velocity motion from time $t$ to $t+1$. This serves as a Gaussian next-state prior centered on a constant velocity motion model, with a cut-off disallowing unreasonable deviations.

This is an instance of a standard chain graphical model~\cite{PictorialStructures}, and we can solve for the best trajectory $\xi^\star = \arg\max_\xi s(\xi)$ efficiently using a max-sum message passing dynamic program.

\out{
We can find $k^* = \arg \max_k S(k)$ using Viterbi decoding on the grid map sequences. We can make decoding more efficient by pruning prefixes with low score, essentially halting message-pass from low-score prefixes in the forward pass. We can also cache the results of $\psi$ in a two-dimensional lookup table to avoid recomputation. 
}

Beyond $\xi^\star$, we are interested in extracting a {\em set} of trajectories which satisfy:
\begin{align}
\{\xi^{\star_1},\ldots, \xi^{\star_k}\}~=~& \underset{ \{\xi^{1},\ldots, \xi^{k}\}}{\arg \max} \sum_{i=1}^k s(\xi^i) \\
\text{subject to:~~} &||\xi^i - \xi^j|| > 1. \label{eq:diversity-norm}
\end{align}
This seeks to find a set of $k$ trajectories which maximizes $s(\cdot)$ but are sufficiently far from each other, for some norm $||\cdot||$. Following~\cite{NBest}, we solve this by iteratively extracting trajectories by solving $s(\xi)$ and then masking regions of $g_t$ to guarantee the distance constraint on the next optimization of $s(\xi)$ for the next trajectory.

 Note this framework is employed only during inference.  Incorporating it into a learned end-to-end system is an interesting avenue for future work.

\subsection{Model}
We employ an encoder-decoder architecture for modelling, where the encoder maps the 4D input tensor (time $\times $ space $\times$ channels) into some internal latent representation, and the decoder uses that representation to model the output distribution over states at a pre-determined set of future time offsets.

\mypar{Encoder:} We use a convolutional network (CNN) backbone of 2D convolutions similar to VGG16~\cite{VGG}, on each 3D tensor of the input sequence. Following~\cite{TCN}, we found that temporal convolutions acheived better performance and significantly faster training than a recurrent neural network (RNN) structure. To incorporate the temporal dimension, we add in two 3D convolutions -- one towards the beginning of the backbone, and one at the end -- of kernel size $3 \times 3 \times 3$ and $4 \times 3 \times 3$ without padding, respectively.

\mypar{Decoder:} We experiment with two different decoding architectures: (1) ``one-shot'' prediction of the entire output sequence, and (2) an RNN-decoder which emits a distribution at each inference recurrence step.

One-shot prediction simply requires a two-layer network to regress all the distribution parameters at once, or a 2D convolutional-transpose network with channels equal to the sequence length. For our RNN-decoder, we use only a single GRU cell, whose hidden output is used to regress the true output, which is then fed in as next input.

For the occupancy grid map output representation, the semantic road map $R$ is fed through a separate, shallow CNN tower ($16 \to 16 \to 1$ filters), yielding a spatial grid. This grid intuitively acts as a prior heatmap of static information that is appended to the decoder before applying softmax. This should allow the model to easily penalize positions corresponding to obstacles and non-drivable surfaces. See Fig. \ref{fig:architectures} for a depiction of the architecture.

\medskip
\subsection{Implementation Details}
\mypar{Spatio-temporal dimensions:} We chose $\ell = 2.5$s of past history, and predict up to $m = 5$s in the future. To reduce the problem space complexity, we also subsample the past at $2$ Hz and predict future control points at $1$s intervals. 
We chose the input frames to be $128 \times 128$ pixels, and chose the corresponding real-world extents to be $50 \times 50 m^2$. The output grid size was set to $64 \times 64$ pixels, so that each pixel covers $0.78 m^2$.
For sampling a diverse trajectory set, we choose the norm in Equation~\ref{eq:diversity-norm} to be $||\text{diag}([0,1/3,1/3,1/5,1/5])x||_\infty$, in pixel coordinates, which forces greater spatial diversity in later time steps between trajectories.

\mypar{Model:} To conserve memory in our CNN backbone, we use separable 2D convolutions for most of the backbone with stride 2 (see Fig.~\ref{fig:architectures}). We employ batch-normalization layers to deal with the different dynamic ranges of our input channels (\eg, binary masks, RGB pixel values, $m/s$, $m/s^2$). We use CoordConv~\cite{Coordconv} for all 2D convolutions in the encoder, which aids in mapping spatial outputs to regressed coordinates.  We chose a $512$-dimensional latent representation as the final layer of the encoder, and for all GRU units.

\mypar{Training:}
For training, we started with a learning rate of $10^{-4}$ and Adam optimizer, and switched to SGD with a learning rate of $10^{-5}$ after convergence with Adam. For training Gaussian parameters we found that pretraining the model on mean-squared error produced better trajectories. All models were trained on a single NVIDIA Maxwell Titan X GPU with Tensorflow Keras.  On the dataset described next, training models took approximately 14 hours\footnote{Training time was heavily dominated by reading data from disk, which could be significantly optimized with more efficient I/O strategies.}.

%% file: experiments.tex
\input{related-datasets-table}

\section{Experiments}

\begin{table*}
\begin{center}
\small
\begin{tabular}{ c || c | c | c | c || c || c  c  c  }
 \specialrule{.1em}{.05em}{.05em}
\multicolumn{1}{c ||}{}    & 
\multicolumn{4}{c ||}{}   & 
\multicolumn{1}{c ||}{}    & 
\multicolumn{3}{c}{mean L2 \;/\; hit-rate $<1m$} \\
Method & Target & Other & Road & RNN & RMSE  & 1 sec & 2 sec  & 5 sec \\
\hline

Gaussian Regression & \checkmark & & &   & 2.55 & 0.50 / 0.79 & 1.04 / 0.59 & 3.91 / 0.34 \\
 & \checkmark & \checkmark & &  & 2.12 & 0.48 / 0.80 & 0.96 / 0.60 & 3.36 / 0.35 \\
 & \checkmark& \checkmark& \checkmark  & & 1.90 & 0.47 / 0.84 & 0.94 / 0.64 & 3.03 / 0.35 \\
 & \checkmark& \checkmark& \checkmark & \checkmark & \textbf{1.82} & \textbf{0.44} / \textbf{0.88} & \textbf{0.86} / 0.66 & \textbf{2.99} / 0.34 \\
\hline
Grid map Top-1 & \checkmark & &  & & 3.53 & 0.87 / 0.63 & 1.31 / 0.52 & 5.04 / 0.42 \\
 & \checkmark & \checkmark & &  &  2.71 & 0.77 / 0.64 & 1.17 / 0.52 & 4.40 / 0.42 \\
 & \checkmark& \checkmark&   \checkmark & & 2.05 & 0.61 / 0.86 & 1.00 / \textbf{0.69} & 3.37 / 0.40 \\
 & \checkmark& \checkmark&   \checkmark & \checkmark & 1.99 & 0.55 / 0.87 & 0.97 / 0.67 & 3.23 / \textbf{0.42} \\
\specialrule{.1em}{.05em}{.05em}
\end{tabular}
\end{center}
\caption{Ablation study on Gaussian Regression trajectories and Grid Map methods.}
\label{tab:ablation}
\end{table*}

\begin{table}[]
\begin{center}
\small
\begin{tabular}{c || c || c  c  c}
\specialrule{.1em}{.05em}{.05em}
Method & RMSE  & 1 sec & 2 sec  & 5 sec \\
\hline
Linear & 3.53 & \textbf{0.37} / 0.89 & 1.08 / 0.62 & 5.87 / 0.26 \\
Industry & 2.31 & \textbf{0.37} / \textbf{0.90} & 0.92 / \textbf{0.67} & 4.18 / 0.40 \\
Gauss. Reg. & \textbf{1.82} & 0.44 / 0.88 & \textbf{0.86} / 0.66 & \textbf{2.99} / 0.34 \\
GMM &  2.33 & 0.48 / 0.82 & 0.95 / 0.65 & 4.17 / 0.19 \\
Grid map &  1.99 & 0.55 / 0.87 & 0.97 / \textbf{0.67} & 3.23 / \textbf{0.42} \\
\specialrule{.1em}{.05em}{.05em}
GMM top-5 &  1.58 & 0.43 / 0.89 & 0.79 / 0.70 & 2.54 / 0.32 \\
Grid top-5 & 1.25 & 0.47 / 0.87 & 0.82 / 0.69 & 1.39 / 0.56 \\
\specialrule{.1em}{.05em}{.05em}
\end{tabular}
\end{center}
\caption{Results for multimodal prediction methods.}
\label{tab:results}
\end{table}

\subsection{Dataset} 
We collected a substantial dataset to evaluate our methods. It is over an order of magnitude larger than~\cite{IntentNet} and many orders of magnitude larger than KITTI~\cite{Kitti}.
Our dataset consists of tracked vehicles in view of an ego data collection vehicle driving in a dense urban area of San Francisco. A state-of-the-art perception and localization stack processes multiple sensor modalities to produce top-down projected 2D bounding boxes in world coordinates, as well as $v_i^t$, $a_i^t$ and $\Sigma_{i\{x,v,a\}}^t$. We also include the high-definition rendering of the road network, with detailed annotations described in Section~\ref{sec:roadgraph}.

The dataset has more than $6.25$ million frames from more than $173$ hours of driving in the period of June--July 2018. We split the data into non-overlapping events of $7.5$ seconds, and extract $72,878$ train and $10,473$ test events.  
Events are collected near intersections to skew the dataset distribution towards non-trivial and non-straight driving behavior.  To measure our methods' generalization power to new intersections, the test/train partition ensures no intersections appear in both; there are $79$ and $9$ unique intersections in train and test, respectively. To reduce the bias of any one geographic location, we capped the number of samples per intersection at $5,000$.  Note that intersections can be much more complex than simple 4-way junctions.

For an overview of related datasets' attributes and sizes, please refer to Table~\ref{tab:datasets}. Note that other vehicle datasets are either quite small (\eg, KITTI~\cite{Kitti}), or are not available for comparison.  The Stanford Drone dataset is comparable in size and provides a natively top-down representation, which allows us to extend our model to pedestrian prediction in the future.

\begin{figure*}
\begin{center}
\subfloat[Gaussian Regression]{{\includegraphics[width=0.5\linewidth,height=4cm]{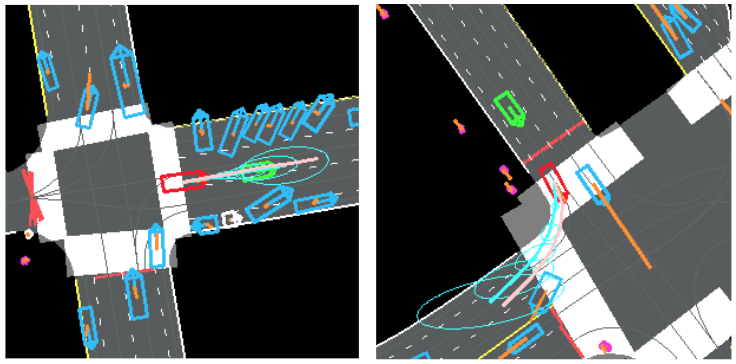} }}%
\subfloat[GMM-CVAE]{{\includegraphics[width=0.5\linewidth,height=4cm]{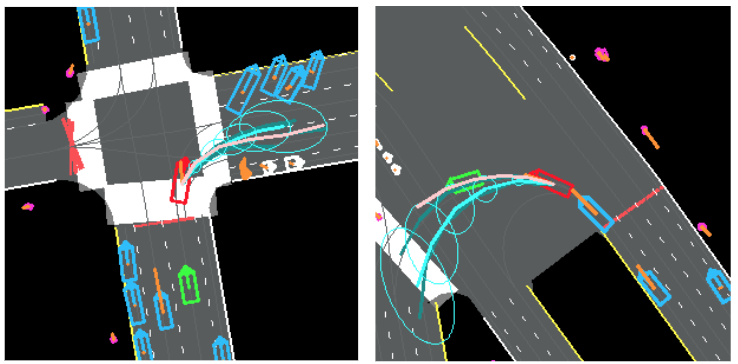} }}%
\vspace{-0.2cm}
\caption{\small Examples of Gaussian Regression and GMM-CVAE methods. Ellipses represent a standard deviation of uncertainty, and are only drawn for the top trajectory; only trajectories with probability $>0.05$ are shown, with cyan the most probable.We see that uncertainty ellipses are larger when turning than straight, and often follow the direction of velocity. In the GMM-CVAE example, different samples result in turning into different lanes in a junction.}%

\label{fig:gt}%
\end{center}
\end{figure*}
\begin{figure*}
\vspace{-0.5cm}
\begin{center}
\includegraphics[width=0.99\linewidth,height=4cm]{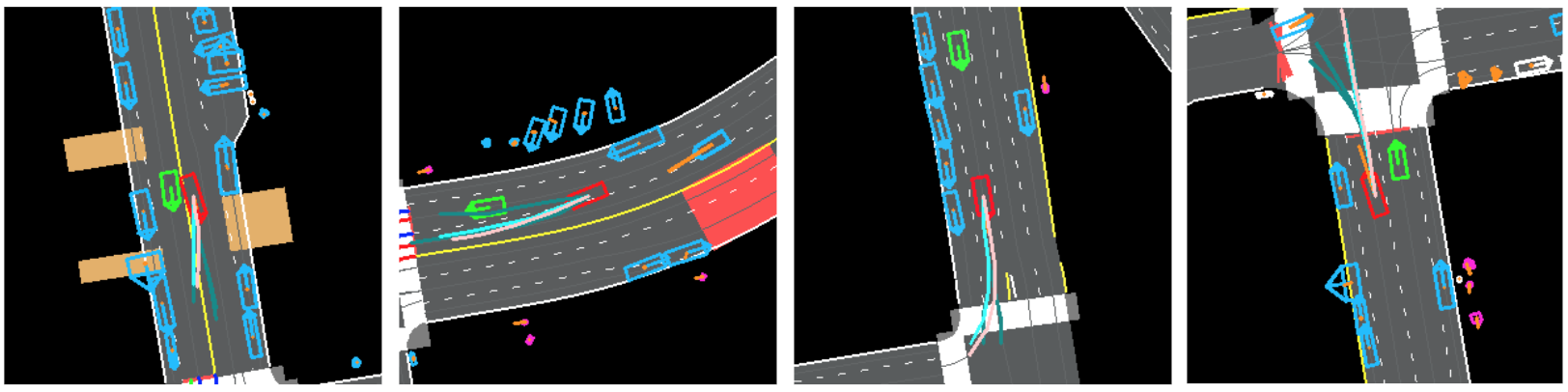}
\end{center}
\vspace{-0.5cm}
\caption{\small Examples of trajectories sampled from the Grid Map method. 
The rightmost example is a failure case, as the method predicts a mode that turns into oncoming traffic; however, such traffic rules may be hard to discern from only a road map. The method predicts sophisticated behavior such as maneuvering around vehicles and changing lanes.
}
\label{fig:gm}%
\vspace{-0.4cm}
\end{figure*}

\subsection{Results}

We measure performance of all methods on root mean-squared error (RMSE) over all future timesteps $\Delta t \in \{1,2,3,4,5\}$ seconds, as well as the following metrics on 1, 2, 5s futures:  (1) L2 distance between the mean predictions and ground truth, (2) hit-rate of L2 distance under a $1 m$  threshold, and (3) oracle error, defined as the minimum L2 distance (\ie $\min_i \norm{Y - \xi^i}_2$) out of top-$k$ predictions.

\mypar{Models:} We compare the following methods:
\begin{enumerate}[nolistsep,noitemsep,wide,labelwidth=!,labelindent=0pt]
    \item[$\cdot$] \textit{Linear}: Baseline that extrapolates trajectories using the most recent velocity and acceleration, held constant.
    \item[$\cdot$] \textit{Industry}: Closed-source industry method used in real-world autonomous vehicle driving.  It consists of a hybrid of physics, hand-designed rules, and targeted machine learning classification models.  Specific problem modes are modeled in a custom fashion, requiring multiple person-years of modeling effort.
    \item[$\cdot$] \textit{Regression with Gaussian Uncertainty}: Our method to regress a Gaussian distribution per future timestep.
    \item[$\cdot$] \textit{Multi-modal Gaussian Regression (GMM-CVAE)}: Our described method to predict a set of Gaussians, sampling from a categorial latent variable.
    \item[$\cdot$] \textit{Grid map}: Our method to predict occupancy grid maps. To compare to the other methods, we extract trajectories using the described trajectory sampling procedure. 
\end{enumerate}

\mypar{Ablation study:} To determine the efficacy of different input channels, we group them and evaluate their performance as follows:
\begin{enumerate}[nolistsep,wide,labelwidth=!,labelindent=0pt]
    \item[$\cdot$] \textit{Target state}: State of the entity of interest including its rendered past and present bounding boxes.
    \item[$\cdot$] \textit{Other state}: Features for other dynamic entities including their bounding boxes rendered.
    \item[$\cdot$] \textit{Road map}: Road map and traffic light rendering.
\end{enumerate}

\out{
As expected, each new feature improved model performance. However, adding in feature for other dynamic entities only provided marginal improvement, suggesting a more clever integration of dynamic interactions is needed for the feature to be useful. 

We also varied whether to use a RNN decoder. Using an RNN improved performance, though only marginally, but sacrifices higher training and inference time to do so. We refer to Table 1 for a summary of the results.
}

\medskip
\noindent As shown in Table~\ref{tab:ablation}, each feature type contributes to improved model performance: Adding the dynamic context of other entities improves 5s prediction by 0.55m in average L2-error, and including the road map adds another 0.33m improvement.  See the supplementary material for qualitative visualizations of how ablated features contribute to performance.

\mypar{Quantitative method comparison:}
In Table ~\ref{tab:results}, we compare all methods using all features. Evaluating the best-in-top-5 trajectory performed better than the single MAP trajectory in all metrics, indicating some value of probabilistic prediction over multiple modes.

In general our mixture of sampled Gaussian trajectories underperformed our other proposed methods; we observed that some samples were implausible. We leave it to future work to determine better techniques for obtaining a diverse set of trajectory samples directly from a learned model.

Interestingly, both Linear and Industry baselines performed worse relative to our methods at larger time offsets, but better better at smaller offsets. This can be attributed to the fact that predicting near futures can be accurately achieved with classical physics (which both baselines leverage)---more distant future predictions, however, require more challenging semantic understanding.

Note that while all models are evaluated here in terms of L2-error, none of the models directly optimize this quantity but instead optimize the likelihood of distributions over the future state space, which has other benefits over regression---this is demonstrated in the top-5 metric, as well as in the qualitative results below.

\mypar{Qualitative Analysis:} 
We show examples of the Gaussian Regression and GMM-CVAE trajectories in Fig.~\ref{fig:gt}, and sampled Grid Map trajectories in Fig.~\ref{fig:gm}. Please see the supplementary material for more examples and visualizations.
Overall, trajectories have plausibly learned traffic rules: lane-keeping, traffic light obeyance, following behavior, and even the illegal ones are specious.

\out{
  Youtube account:
  topdownpredictioncvpr2019@gmail.com 
  password is: rulesoftheroad
}

\section{Conclusion}
We present a unified framework for multimodal future state prediction. Our novel input encoding encapsulates static and dynamic scene context, taking advantage of measurements from sensor modalities and a high-definition road map. We experiment with continuous and discrete output representations, and arrive at solutions that address the uncertainty and multi-modality of future prediction.
Empirical and qualitative evaluation show that our methods improve on baselines that do not encode scene context, and  successfully create diverse samples in complex driving scenarios. 

\out{
There are several avenues for future work: (1) augmenting the input representation with additional signals (vehicle lights; human pose, gestures or appearance), (2) pedestrian modeling, (3) learning more effective multi-modal parametric distributions and (4) incorporating trajectory sampling from grid maps (as in Sec.~\ref{sec:gm}) into the learning process.
}

%% file: related-datasets-table.tex
\begin{table}[]
\footnotesize
\begin{center}
\begin{tabular}{lllcrr}
 \specialrule{.1em}{.05em}{.05em}
Dataset  & Methods & \shortstack{\\Mode \\ type} & \shortstack{Road \\ info} & \shortstack{\# tracks \\ $>3s$} & \shortstack{\# \\scenes} \\
\hline
\shortstack{\\ ETH+UCY}         & SocialLSTM$^*$~\cite{Sociallstm}  & Peds  & no         & 1,536           & 4   \\
Stanford Drone      & DESIRE~\cite{Desire}  & Peds   & no      & 19,564         & 100    \\
KITTI           & DESIRE~\cite{Desire}  & Cars   & no       & 309         & 20  \\
CaliForecasting$^\dagger$ & R2P2~\cite{R2p2}, C3PO$^\dagger$ & Cars   & no       & 10,000         & ---   \\
Ours        & Ours & Cars   & yes      & 72,878         & 79    \\
\hdashline[1pt/1pt]
\shortstack{\\ {\em private }} & FaF~\cite{Faf} & Cars & no & --- & --- \\
{\em private } & IntentNet~\cite{IntentNet} & Cars & yes & --- & --- \\
{\em private } & ChauffeurNet$^\dagger$ & Cars & yes & --- & --- \\
\specialrule{.1em}{.05em}{.05em}

\multicolumn{3}{l}{$*$: Code is publicly available.} \\
\multicolumn{3}{l}{$\dagger$: Method or dataset not yet published.}
\end{tabular}
\end{center}
\caption{Overview of related public datasets' {\em training set} statistics.}
\label{tab:datasets}
\vspace{-0.5cm}
\end{table}

%% file: paper.bbl
\begin{thebibliography}{10}\itemsep=-1pt

\bibitem{Sociallstm}
A.~Alahi, K.~Goel, V.~Ramanathan, A.~Robicquet, L.~Fei-Fei, and S.~Savarese.
\newblock {SocialLSTM}: Human trajectory prediction in crowded spaces.
\newblock {\em CVPR}, 2016.

\bibitem{bagautdinov2017social}
T.~M. Bagautdinov, A.~Alahi, F.~Fleuret, P.~Fua, and S.~Savarese.
\newblock Social scene understanding: End-to-end multi-person action
  localization and collective activity recognition.
\newblock In {\em CVPR}, 2017.

\bibitem{TCN}
S.~Bai, J.~Z. Kolter, and V.~Koltun.
\newblock An empirical evaluation of generic convolutional and recurrent
  networks for sequence modeling.
\newblock {\em arXiv:1803.01271}, 2018.

\bibitem{Chauffernet}
M.~Bansal, A.~Krizhevsky, and A.~Ogale.
\newblock Chauffeurnet: Learning to drive by imitating the best and
  synthesizing the worst.
\newblock {\em arXiv preprint arXiv:1812.03079}.

\bibitem{Bimodal}
J.~Behboodian.
\newblock On the modes of a mixture of two normal distributions.
\newblock {\em Technometrics}, pages 131--139, 1970.

\bibitem{ego-pred}
A.~Bhattacharyya, M.~Fritz, and B.~Schiele.
\newblock Long-term on-board prediction of people in traffic scenes under
  uncertainty.
\newblock In {\em CVPR}, 2018.

\bibitem{LDA}
D.~M. Blei, A.~Y. Ng, and M.~I. Jordan.
\newblock Latent dirichlet allocation.
\newblock {\em JMLR}, 2003.

\bibitem{nvidia-sdc}
M.~Bojarski, D.~Del~Testa, D.~Dworakowski, B.~Firner, B.~Flepp, P.~Goyal, L.~D.
  Jackel, M.~Monfort, U.~Muller, J.~Zhang, et~al.
\newblock End to end learning for self-driving cars.
\newblock {\em arXiv preprint arXiv:1604.07316}, 2016.

\bibitem{Bullinger_2018_ECCV}
S.~Bullinger, C.~Bodensteiner, M.~Arens, and R.~Stiefelhagen.
\newblock 3d vehicle trajectory reconstruction in monocular video data using
  environment structure constraints.
\newblock In {\em ECCV}, 2018.

\bibitem{IntentNet}
S.~Casas, W.~Luo, and R.~Urtasun.
\newblock Intentnet: Learning to predict intention from raw sensor data.
\newblock In {\em CoRL}, 2018.

\bibitem{AutoX}
C.~Chen, A.~Seff, A.~Kornhauser, and J.~Xiao.
\newblock Deepdriving: Learning affordance for direct perception in autonomous
  driving.
\newblock In {\em ICCV}, 2015.

\bibitem{BaiduBEV}
X.~Chen, H.~Ma, J.~Wan, B.~Li, and T.~Xia.
\newblock Multi-view 3d object detection network for autonomous driving.
\newblock In {\em CVPR}, 2017.

\bibitem{CarFusion}
N.~Dinesh~Reddy, M.~Vo, and S.~G. Narasimhan.
\newblock Carfusion: Combining point tracking and part detection for dynamic 3d
  reconstruction of vehicles.
\newblock In {\em CVPR}, 2018.

\bibitem{PictorialStructures}
P.~F. Felzenszwalb and D.~P. Huttenlocher.
\newblock Pictorial structures for object recognition.
\newblock {\em IJCV}, 61(1):55--79, 2005.

\bibitem{Kitti}
A.~Geiger, P.~Lenz, and R.~Urtasun.
\newblock Are we ready for autonomous driving? the kitti vision benchmark
  suite.
\newblock {\em CVPR}, 2012.

\bibitem{Multi-human}
B.~Ivanovic, E.~Schmerling, K.~Leung, and M.~Pavone.
\newblock Generative modeling of multimodal multi-human behavior.
\newblock 2018.

\bibitem{GumbelSoftmax}
E.~Jang, S.~Gu, and B.~Poole.
\newblock Categorical reparameterization with gumbel-softmax.
\newblock {\em ICLR}, 2017.

\bibitem{Uncertainty}
A.~Kendall and Y.~Gal.
\newblock What uncertainties do we need in bayesian deep learning for computer
  vision?
\newblock {\em NIPS}, 2017.

\bibitem{Kitani}
K.~M. Kitani, B.~D. Ziebart, J.~A. Bagnell, and M.~Hebert.
\newblock Activity forecasting.
\newblock {\em ECCV}, 2012.

\bibitem{ActivityPred}
Y.~Kong and Y.~Fu.
\newblock Human action recognition and prediction: A survey.
\newblock {\em arXiv preprint arXiv:1806.11230}, 2018.

\bibitem{Kooij2014ContextBasedPP}
J.~F.~P. Kooij, N.~Schneider, F.~Flohr, and D.~Gavrila.
\newblock Context-based pedestrian path prediction.
\newblock In {\em ECCV}, 2014.

\bibitem{Desire}
N.~Lee, W.~Choi, P.~Vernaza, C.~B. Choy, P.~H.~S. Torr, and M.~K. Chandraker.
\newblock {DESIRE:} distant future prediction in dynamic scenes with
  interacting agents.
\newblock {\em CVPR}, 2017.

\bibitem{Coordconv}
R.~Liu, J.~Lehman, P.~Molino, F.~P. Such, E.~Frank, A.~Sergeev, and
  J.~Yosinski.
\newblock An intriguing failing of convolutional neural networks and the
  {CoordConv} solution.
\newblock {\em arXiv preprint arXiv:1807.03247}, 2018.

\bibitem{VideoPred}
W.~Lotter, G.~Kreiman, and D.~D. Cox.
\newblock Deep predictive coding networks for video prediction and unsupervised
  learning.
\newblock {\em CoRR}, 2016.

\bibitem{Faf}
W.~Luo, B.~Yang, and R.~Urtasun.
\newblock Fast and furious: Real time end-to-end 3d detection, tracking and
  motion forecasting with a single convolutional net.
\newblock {\em CVPR}, 2018.

\bibitem{mousavian20173d}
A.~Mousavian, D.~Anguelov, J.~Flynn, and J.~Ko{\v{s}}eck{\'a}.
\newblock 3d bounding box estimation using deep learning and geometry.
\newblock In {\em CVPR}, 2017.

\bibitem{NBest}
D.~Park and D.~Ramanan.
\newblock N-best maximal decoders for part models.
\newblock {\em ICCV}, 2011.

\bibitem{ALVINN}
D.~A. Pomerleau.
\newblock Alvinn: An autonomous land vehicle in a neural network.
\newblock In {\em NIPS}, 1989.

\bibitem{PointNet}
C.~R. Qi, H.~Su, K.~Mo, and L.~J. Guibas.
\newblock Pointnet: Deep learning on point sets for 3d classification and
  segmentation.
\newblock {\em CoRR}, 2016.

\bibitem{ego-pred2}
N.~Rhinehart and K.~M. Kitani.
\newblock First-person activity forecasting with online inverse reinforcement
  learning.
\newblock In {\em ICCV}, 2017.

\bibitem{R2p2}
N.~Rhinehart, K.~M. Kitani, and P.~Vernaza.
\newblock R2p2: A reparameterized pushforward policy for diverse, precise
  generative path forecasting.
\newblock {\em ECCV}, 2018.

\bibitem{VGG}
K.~Simonyan and A.~Zisserman.
\newblock Very deep convolutional networks for large-scale image recognition.
\newblock {\em CoRR}, 2014.

\bibitem{Gridmap}
S.~Thrun, W.~Burgard, and D.~Fox.
\newblock {\em Probabilistic Robotics}.
\newblock MIT Press, 2005.

\bibitem{Trajectory-gmm}
J.~Wiest, M.~Hoffken, U.~Kresel, and K.~Dietmayer.
\newblock Probabilistic trajectory prediction with gaussian mixture models.
\newblock {\em Intelligent Vehicles Symposium}, 2012.

\bibitem{VoxelNet}
Y.~Zhou and O.~Tuzel.
\newblock {VoxelNet}: End-to-end learning for point cloud based 3d object
  detection.
\newblock {\em CoRR}, 2017.

\end{thebibliography}
